\documentclass[letterpaper]{article} 
\usepackage{aaai2026}  
\usepackage{times}  
\usepackage{amsmath} 
\usepackage{amsfonts} 
\usepackage{pifont} 
\usepackage{multirow} 
\usepackage{helvet}  
\usepackage{courier}  
\usepackage[hyphens]{url}  
\usepackage{graphicx} 
\usepackage{natbib}  
\usepackage{caption} 
\usepackage{algorithm}
\usepackage{algorithmic}

\usepackage{newfloat}
\usepackage{listings}
\DeclareCaptionStyle{ruled}{labelfont=normalfont,labelsep=colon,strut=off} 
\floatstyle{ruled}
\newfloat{listing}{tb}{lst}{}
\floatname{listing}{Listing}
\nocopyright 

\title{To Align or Not to Align: Strategic Multimodal Representation Alignment for
Optimal Performance}
\author{
Wanlong Fang\textsuperscript{\rm 1,2},
Tianle Zhang\textsuperscript{\rm 2},
Alvin Chan\textsuperscript{\rm 2,3}\thanks{Corresponding Author.}
}
\affiliations{
\textsuperscript{\rm 1}Artificial Intelligence-X (AI-X) @ NTU, Interdisciplinary Graduate Programme, Nanyang Technological University\\
\textsuperscript{\rm 2}College of Computing and Data Science, Nanyang Technological University\\
\textsuperscript{\rm 3}Lee Kong Chian School of Medicine, Nanyang Technological University\\
\{wanlong001, tianle004\}@e.ntu.edu.sg,\; guoweialvin.chan@ntu.edu.sg
}

\usepackage{bibentry}

\begin{document}

\maketitle

\begin{abstract}
Multimodal learning often relies on aligning representations across modalities to enable effective information integration—an approach traditionally assumed to be universally beneficial. However, prior research has primarily taken an observational approach, examining naturally occurring alignment in multimodal data and exploring its correlation with model performance, without systematically studying the direct effects of explicitly enforced alignment between representations of different modalities. In this work, we investigate how explicit alignment influences both model performance and representation alignment under different modality-specific information structures. Specifically, we introduce a controllable contrastive learning module that enables precise manipulation of alignment strength during training, allowing us to explore when explicit alignment improves or hinders performance. 
Our results on synthetic and real datasets under different data characteristics show that the impact of explicit alignment on the performance of unimodal models is related to the characteristics of the data: the optimal level of alignment depends on the amount of redundancy between the different modalities. We can find an optimal alignment strength that balances modality-specific signals and shared redundancy in the mixed information distributions. This work can help practitioners on when and how to enforce alignment for optimal unimodal encoder performance.
\end{abstract}

\section{Introduction}

\begin{figure*}[t]
\centering
\label{Fig1}
    \includegraphics[width=0.9\textwidth]{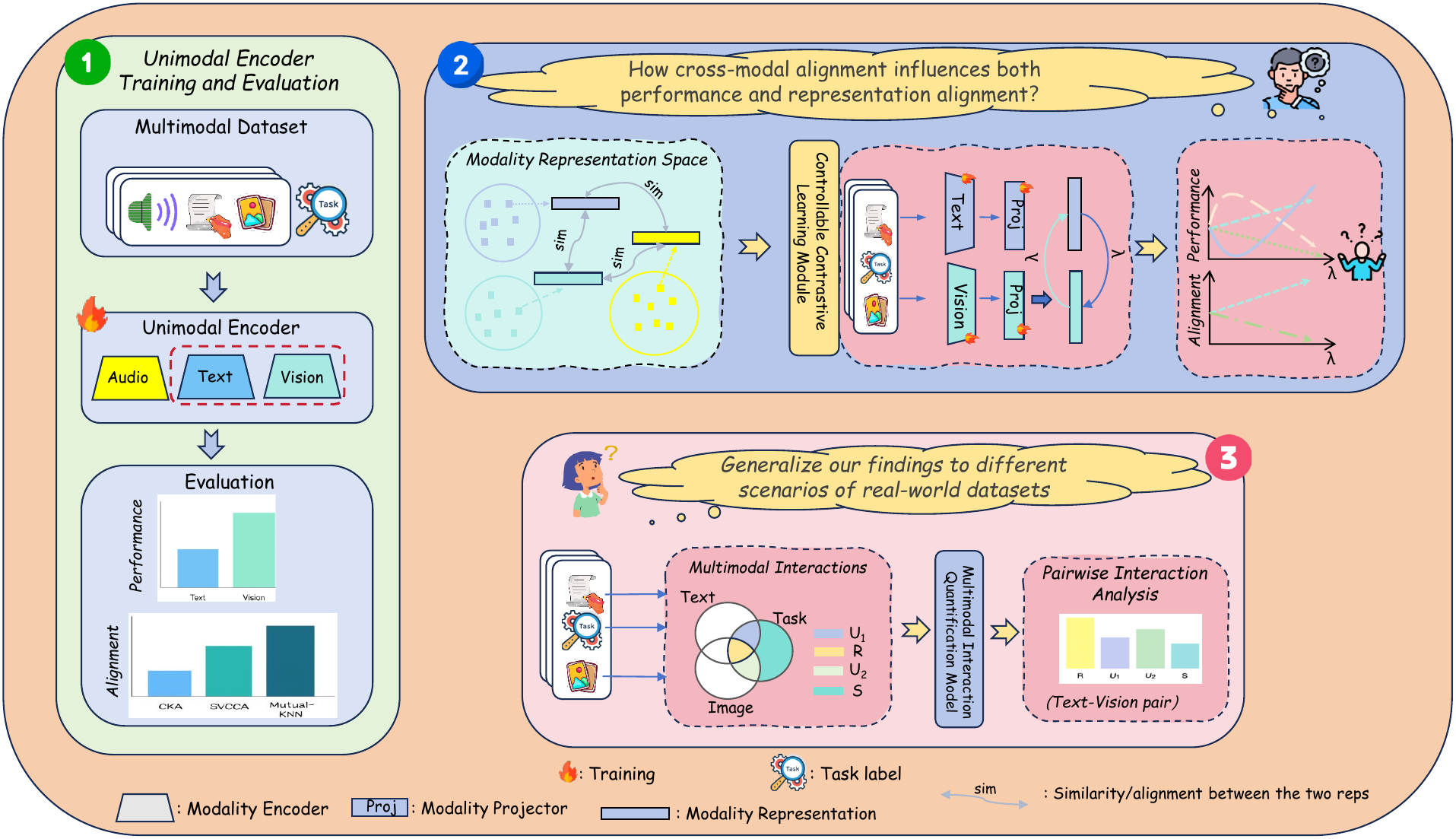}  
\caption{\small  
Overview of our experimental framework for analyzing the impact of explicit multimodal alignment.  
\ding{172} \textbf{Independent Training Baseline:} Each unimodal encoder is trained independently to establish a performance and representation baseline, capturing naturally emergent alignment.  
\ding{173} \textbf{Controlled Cross-modal Alignment:} We introduce a contrastive learning module to explicitly enforce alignment during training, enabling systematic analysis of how alignment strength affects unimodal performance and representational similarity.  
\ding{174} \textbf{Real-world Generalization via PID:} We apply Partial Information Decomposition (PID) to quantify the redundancy–uniqueness–synergy structure across modality pairs, validating our findings under diverse real-world data conditions.}
\end{figure*}

A foundational principle in multimodal machine learning is that explicitly enforcing alignment between modalities in a shared semantic space is essential for effective knowledge fusion and improved performance~\cite{wang2020deep,zadeh-etal-2017-tensor,jin2025multimodal, zong2024self,wang-shi-2023-video,wang-etal-2023-balance}. This assumption has driven the development of state-of-the-art architectures, which often use contrastive learning objectives to maximize representation similarity~\cite{yuan2021multimodal, pielawski2020comir}.

However, recent work shows that representation alignment can also emerge naturally from data itself without a direct alignment objective. The Platonic Representation Hypothesis~\cite{huh2024platonicrepresentationhypothesis} posits that as unimodal models increase in scale and capability, their internal representations naturally converge toward a shared statistical model of the underlying reality. This emergent alignment is not coincidental; it correlates with the model's performance. While the Platonic Representation Hypothesis suggests a universal benefit, critical empirical findings from \cite{tjandrasuwita2025understandingemergencemultimodalrepresentation} demonstrate that the utility of alignment is far from universal. Their work reveals that the relationship between alignment and performance is highly conditional on the data's intrinsic information structure. 

Prior research has primarily taken an observational approach, examining natural alignment in multimodal data and exploring its correlation with model performance. However, no studies have previously systematically varied the strength of alignment enforcement to assess its effects on model behavior. Gaining a clearer understanding of this area could offer valuable insights into how alignment influences both performance and representational structure, which are central considerations in the design of multimodal systems.

In this paper, we take a step toward addressing this gap by exploring the relationship between explicit alignment strength and its effects on model performance and representational similarity. We examine this relationship across different underlying information structures in the data, with the goal of better understanding how alignment may function under varying conditions. Our main research questions are as follows:
\begin{itemize}
\item \textbf{\textit{RQ1}}: Under what conditions does explicit alignment improve or hinder unimodal performance?
\item \textbf{\textit{RQ2}}: How does this improvement generalize to real-world multimodal tasks?
\end{itemize}
To systematically investigate these questions, we introduce a controllable contrastive alignment loss weighted by a scalar factor $\lambda$, which is added to the task loss during unimodal encoder training. This allows us to precisely adjust alignment strength and evaluate its effects on both performance and representational similarity between modalities.
In the synthetic dataset setting, we explicitly manipulate the degree of redundancy, enabling controlled testing under known information distributions. These experiments showed a direct link between representation alignment and improvement of encoders when redundancy is present between the modalities. To validate these findings in practical applications, we first quantify the level of redundant, unique, and synergistic information across modality pairs in real-world datasets, using the Partial Information Decomposition (PID) framework \cite{williams2010nonnegative,bertschinger2014quantifying}. We then conduct experiments across all possible modality pairs within three diverse real-world benchmarks, offering a comprehensive analysis of alignment utility under varying data conditions. We also found that our approach to enforce alignment between modalities improve the performance of unimodal encoders, making it a practical tool for unimodal tasks. 

Our findings provide compelling empirical evidence regarding the conditional utility of explicit alignment, making the following key contributions and impacts:
\begin{itemize}
    \item \textbf{\textit{A New Angle on Representation Alignment}}: Unlike prior work, which often investigates implicit alignment, i.e. how representations naturally align within multimodal data. Specifically, we probe the direct link between forced representation alignment and its effects on both empirical performance and representation similarity. 
    \item \textbf{\textit{Findings on Improving Modality Encoders}}: Our experimental results on synthetic datasets show that explicit cross-modal alignment can be a feasible approach to improve unimodal encoders. This improvement depends on the level of redundant information shared between the two modalities.
    
    \item \textbf{\textit{Generalization to Real-world Datasets:}} By applying the PID framework, we validate our findings across real-world multimodal data under different modality-specific information structures, ensuring that insights from synthetic experiments are consistent and robust in practical, non-synthetic scenarios. This cross-validation strengthens the generalization of our findings and demonstrates the applicability of our analysis to real-world tasks.
\end{itemize}

\section{Related works}
\paragraph{Multimodal Representation Alignment.} Prior work has shown that alignment between modality-specific representations can emerge naturally without explicit supervision~\cite{bonheme2022variationalautoencoderslearninsights, huh2024platonicrepresentationhypothesis}, a phenomenon formalized by the \textit{Platonic Representation Hypothesis}~\cite{ziyin2025proofperfectplatonicrepresentation}. Metrics such as CKA~\cite{kornblith2019similarityneuralnetworkrepresentations} and SVCCA~\cite{raghu2017svccasingularvectorcanonical} have been used to study such emergent alignment, which is often correlated with performance gains. However, these studies are mostly observational and do not systematically intervene on alignment strength. Tjandrasuwita et al.~\cite{tjandrasuwita2025understandingemergencemultimodalrepresentation} relate emergent alignment to information structure, but do not explicitly manipulate the alignment mechanism.
\paragraph{Cross-modal alignment via contrastive mearning}
Contrastive learning has become the standard approach for enforcing cross-modal alignment~\cite{radford2021learning, xu2021videoclip, zolfaghari2021crossclr,wangDeepUnifiedCrossModality2021,wang-etal-2025-dream}. While effective, most methods assume that stronger alignment is always beneficial~\cite{cai2025valuecrossmodalmisalignmentmultimodal, jiang2023visionlanguagepretrainingcontrastive}, potentially ignoring modality-specific structures such as redundancy and synergy~\cite{dufumier2025alignmultimodalcontrastivelearning}. Our work departs from this assumption by introducing a controllable contrastive framework that explicitly varies alignment strength and measures its effect on unimodal performance under diverse information structures. \textit{A full related work is provided in extended version.}

\section{Background and Preliminaries}
\paragraph{Multimodal Information Quantification}

To clarify the role of alignment in multimodal learning, we begin by analyzing how information is distributed across modalities.  
We consider two input modalities, $X_1$ and $X_2$, and a shared task label $Y$, where each sample $(x_1, x_2, y)$ is drawn from a joint distribution $\mathcal{P}(X_1, X_2, Y)$.  
While our analysis focuses on two modalities for clarity, the methodology naturally extends to more modalities.

The total information that $(X_1, X_2)$ provides about $Y$ is captured by the multivariate mutual information $I(X_1, X_2; Y)$.  
However, conventional mutual information does not distinguish between different types of interactions among modalities.  
To address this, we adopt the Partial Information Decomposition (PID) framework~\cite{williams2010nonnegative,bertschinger2014quantifying}, which decomposes $I(X_1, X_2; Y)$ into four interpretable components:
\begin{equation}
I(X_1, X_2; Y) = R + U_1 + U_2 + S,
\end{equation}
where:
\begin{itemize}
    \item $R$ is the \textbf{redundancy} — information about $Y$ that is shared by both $X_1$ and $X_2$;
    \item $U_1$, $U_2$ are the \textbf{unique information} contributed individually by $X_1$ and $X_2$, respectively;
    \item $S$ is the \textbf{synergy} — information that only emerges when both modalities are considered jointly.
\end{itemize}
These components satisfy the following consistency constraints based on standard mutual information identities:
\begin{equation}
I(X_1; Y) = R + U_1, \quad I(X_2; Y) = R + U_2.
\end{equation}
This decomposition provides a way to disentangle how each modality contributes to the task. Recent work~\cite{liang2023quantifying, yang2025efficient} has proposed practical estimators for PID components in real-world datasets, which offers the foundation of our empirical analysis.

\section{Method}
The PID framework reveals the intrinsic, static information structure of a dataset—quantifying the redundant, unique, and synergistic components across modality-pairs. This decomposition enables us to generalize our findings across modality pairs with different information structures and provides actionable guidance on when and how strong alignment should be introduced.  To operationalize this analysis, we introduce a controllable contrastive learning module that enables systematic manipulation of alignment strength.  This allows us to directly examine the effect of enforced cross-modal alignment on both unimodal task performance and representation similarity.

\paragraph{Experimental Framework: A Systematic Approach to Probing Alignment Effects}
when multiple modalities (e.g., $X_1, X_2$) are used to predict the same target label $Y$, the representations learned by independently trained unimodal encoders ($f_1, f_2$) should inherently capture some shared, task-relevant information.
For instance, a speaker's facial expressions, vocal tone, and text all convey cues about the same underlying sentiment, thus sharing redundant content while retaining unique signals. While prior work has relied on emergent or maximum alignment, few studies have systematically treated alignment strength as a controllable variable to investigate its impact on unimodal encoder learning. We propose the following procedure to address this gap (illustrated in Figure 1):


\begin{enumerate}
    \item Unimodal encoders are trained independently to establish a baseline.
    \item Experiment on how crossmodal alignment influence both performance and representation alignment.
    \item Generalization to real-world dataset under different data characteristics scenarios.
\end{enumerate}
\paragraph{Controllable Contrastive Learning Module}
\noindent To implement this controllable alignment in practice, we designed a module based on contrastive learning. We consider a batch of $N$ paired samples $\left\{ (x_i^A, x_i^B) \right\}_{i=1}^N$, where $x_i^A$ and $x_i^B$ are inputs from modality $A$ (e.g., visual) and modality $B$ (e.g., textual), respectively. Each input is processed by a corresponding unimodal encoder, $f_A(\cdot)$ or $f_B(\cdot)$, to extract modality-specific features. Following standard practice, these features are projected into a shared, normalized latent space—if dimensional alignment is needed—using separate multi-layer perceptron heads, $g_A(\cdot)$ and $g_B(\cdot)$. The resulting representations are $\mathbf{z}_i^A = g_A(f_A(x_i^A))$ and $\mathbf{z}_i^B = g_B(f_B(x_i^B))$. 

Our symmetric contrastive loss, $\mathcal{L}_{\text{align}}$, is defined as the average of two InfoNCE \cite{oord2018representation} losses, promoting bidirectional alignment. The loss from modality A to modality B, denoted as $\mathcal{L}_{A \to B}$, is formulated as:
\begin{equation}
\mathcal{L}_{A \to B} = -\frac{1}{N} \sum_{i=1}^{N} \log \frac{\exp(\text{sim}(\mathbf{z}_i^A, \mathbf{z}_i^B) / \tau)}{\sum_{j=1}^{N} \exp(\text{sim}(\mathbf{z}_i^A, \mathbf{z}_j^B) / \tau)}.
\end{equation}
where $\text{sim}(\mathbf{u}, \mathbf{v}) = \mathbf{u}^\top \mathbf{v}$ denotes the cosine similarity between two L2-normalized embedding vectors (i.e., their dot product), and $\tau$ is a fixed temperature hyperparameter that controls the sharpness of the softmax distribution. This loss encourages alignment between each positive pair $(\mathbf{z}_i^A, \mathbf{z}_i^B)$ while contrasting it against the $N - 1$ negative samples within the batch.
The complete symmetric alignment loss, $\mathcal{L}_{\text{align}}$, is the average of the two directional losses:
\begin{equation}
\mathcal{L}_{\text{align}} = \frac{1}{2}(\mathcal{L}_{A \to B} + \mathcal{L}_{B \to A}) \quad 
\end{equation}
Finally, this alignment loss is integrated as a regularization term into our model's total training objective:
\begin{equation}
\mathcal{L}_{\text{total}} = \mathcal{L}_{\text{task}} + \lambda \cdot \mathcal{L}_{\text{align}} \quad 
\end{equation}

In this formulation, $\mathcal{L}_{\text{task}}$ denotes the primary loss for the downstream task (e.g., cross-entropy for classification or L1 loss for regression), while $\lambda$ is a scalar hyperparameter controlling the strength of the alignment regularization.
Varying $\lambda$ allows us to systematically explore how explicit alignment influences unimodal encoder performance under different information structures.  
When $\lambda = 0$, the model reduces to independently trained unimodal encoders, serving as the experimental baseline.  
As $\lambda$ increases, the encoders are increasingly regularized to produce similar representations in the latent space.
This formulation enables us to trace unimodal performance across the full spectrum—from no enforced alignment to strong alignment—and positions $\lambda$ as the central experimental lever for addressing our primary research question (RQ1): \textit{Under what conditions does explicit alignment improve or hinder unimodal performance?}

\section{Experimental Setup}
In this section, we describe the experimental setup for both synthetic and real-world datasets. (\textit{Detailed dataset descriptions and training hyperparameters for encoders and modality-specific projectors can be found in the extended version.})

\subsection{Experimental setup on synthetic datasets}
To efficiently investigate our research question, we adopt a controlled experimental setup using synthetic datasets. This approach enables us to \textbf{\textit{systematically manipulate}} the intrinsic properties of multimodal data, particularly the proportions of redundant and unique information.
\begin{figure}
    \centering
    \includegraphics[width=0.86\linewidth]{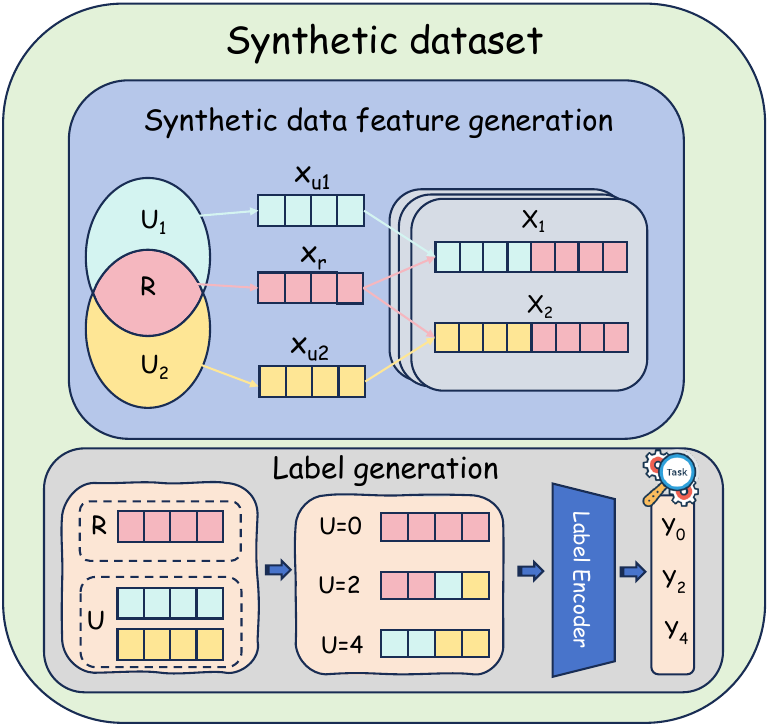}
   \caption{\small Synthetic Data Generation Process.}
    \label{fig:enter-label}
\end{figure}

\subsubsection{Synthetic Data Generation}
We adopt the synthetic dataset proposed by~\cite{tjandrasuwita2025understandingemergencemultimodalrepresentation}, which is specifically designed to construct multimodal inputs $(X_1, X_2)$ and an associated label $Y$ with precisely controlled information composition.  
Following the PID framework, the task-relevant information for $Y$ is decomposed into three conceptual components: a redundant component $x_r$ shared across modalities, and two modality-specific unique components $x_{u1}$ and $x_{u2}$.  
Each modality input is then formed by concatenating the shared and unique components:  \[
x_1 = [x_r, x_{u1}], \quad x_2 = [x_r, x_{u2}].\]  
An illustration of this generation process is provided in Figure~\ref{fig:enter-label}.
The degrees of redundancy ($R$) and uniqueness ($U_i$) are governed by the construction of the label $Y$, which is defined as a non-linear function over a selected subset $Q \subseteq \{x_r, x_{u1}, x_{u2}\}$.  
Specifically, $R$ corresponds to the number of features in $Q$ drawn from $x_r$, and $U_i$ corresponds to the number of features drawn from $x_{ui}$.  
The total uniqueness is defined as $U = |Q| - R$.
By fixing the total number of task-relevant features $|Q|$ and varying the allocation between redundant and unique components, we generate datasets that span a continuum of information structures—from fully redundant to fully modality-specific.  
This controlled design enables systematic investigation of how the alignment-performance relationship varies under different redundancy–uniqueness trade-offs.

\subsubsection{Model Architecture and Training Protocol}

In our synthetic experiments, we use MLPs as unimodal encoders $f_A$ and $f_B$, trained jointly with the controllable contrastive learning module. To ensure consistency and maintain a lightweight setup, each MLP has a fixed hidden dimension of 12—comprising 8 dimensions for shared features and 4 for unique features. 

We evaluate model performance and compute alignment between the unimodal encoders $f_A$ and $f_B$, each trained on $X_1$ and $X_2$, respectively, under varying levels of unique information and different alignment strengths $\lambda \in \{0, 0.2, 0.4, 0.6, 0.8, 1, 2\}$.

\subsection{Experimental setup on real-world datasets}
To generalize our findings from synthetic dataset, we conduct analogous experiments on a selection of real-world multimodal datasets.

\subsubsection{Dataset Selection and Characterization}
We select three representative datasets from \textbf{MultiBench} \cite{liang2021multibenchmultiscalebenchmarksmultimodal}, each exhibiting distinct information characteristics quantified using the PID framework (\textbf{\textit{Table}} \ref{tab:pid}): \textbf{CMU-MOSEI}~\cite{zadeh2018multimodal}, \textbf{AV-MNIST}~\cite{perez2019mfas} and \textbf{MUSTARD}~\cite{castro2019towards}.
These datasets cover a spectrum of redundant, unique, and synergistic information compositions, serving as a diverse testbed for evaluating the impact of alignment. 

\subsubsection{Model Architecture and Training Protocol}
For the affective computing datasets (CMU-MOSEI and MUStARD), we use transformer encoders trained on pre-extracted video, audio, and text features.  
For AV-MNIST, we adopt a vision transformer for digit images and a separate transformer for audio inputs, following the same architecture.
Our training protocol mirrors that of the synthetic experiments.  
For datasets with more than two modalities (e.g., CMU-MOSEI, MUStARD), we conduct experiments on all pairwise combinations (e.g., Vision–Audio, Vision–Text, Audio–Text).  
For each modality pair, we learn dedicated projection heads and apply our controllable contrastive alignment module to regularize cross-modal representations.
To study the effect of alignment strength, we vary the alignment weight $\lambda$ over the range $\{0, 0.1, 0.25, 0.5, 0.6, 0.75, 1, 2, 4\}$ and evaluate unimodal task performance and representational alignment across modalities.

\section{Results and Analysis}

\begin{figure*}[t]
    \centering
    \includegraphics[width=0.95\textwidth]{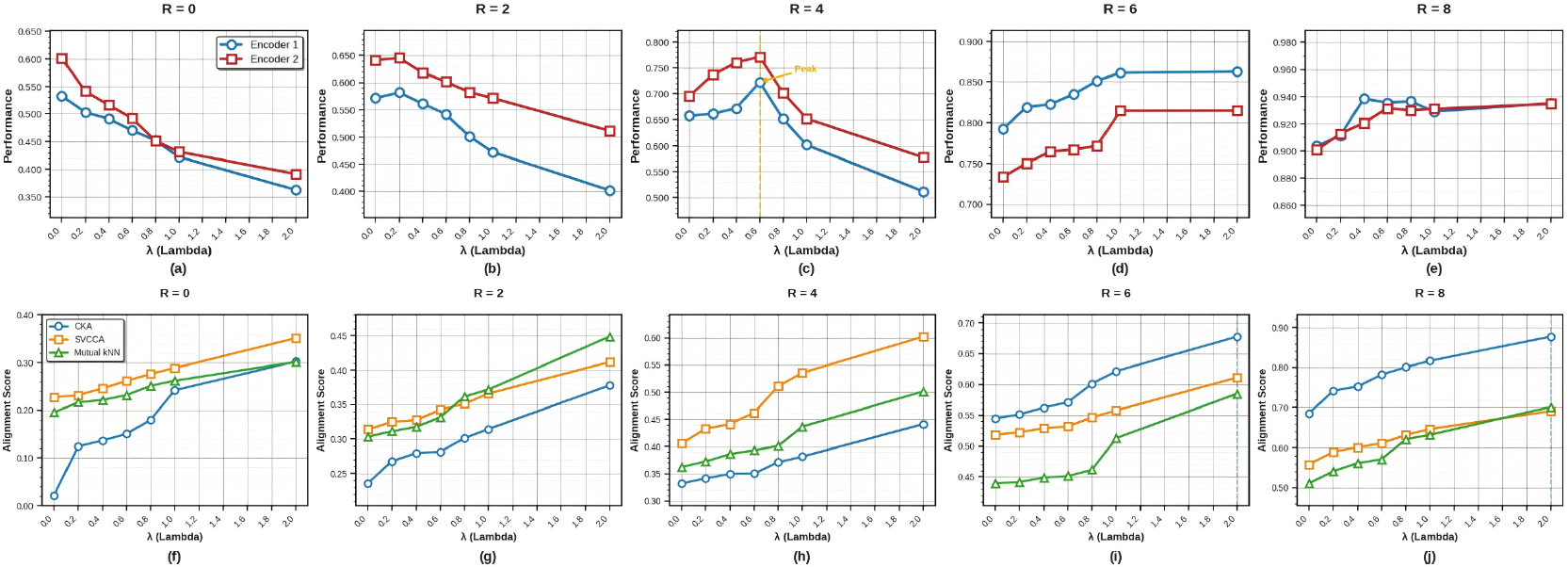}
   \caption{\small  
Experimental results on the synthetic dataset across varying redundancy levels (\textbf{\textit{R}}).  
(a)–(e): Classification performance of Encoder 1 (blue) and Encoder 2 (red) as the alignment strength $\lambda$ increases, under different redundancy levels ($R = 0, 2, 4, 6, 8$).  
(f)–(j): Cross-modal representation alignment scores (CKA, SVCCA, and Mutual-KNN) between the two encoders as a function of $\lambda$, evaluated at the same redundancy levels.
}
    \label{figure3}
\end{figure*}

We introduce our evaluation protocols and alignment metrics, followed by results and analysis on synthetic and real-world datasets. For real-world data, we focus on representative modality pairs to illustrate alignment effects. \textit{Due to space constraints, results for additional combinations are provided in the extended version.}

\subsubsection{Evaluation Metrics \& Alignment Computing}
We evaluate task performance using classification accuracy across all experiments, including both synthetic and real-world datasets.  
For MOSEI, we binarize the original regression labels into positive and negative sentiment classes to align with the classification setting.
To quantify cross-modal alignment between unimodal encoders, we adopt three established metrics:  
\textbf{(i)} Centered Kernel Alignment (\textbf{CKA})~\cite{kornblith2019similarityneuralnetworkrepresentations},  
\textbf{(ii)} Singular Vector Canonical Correlation Analysis (\textbf{SVCCA})~\cite{raghu2017svccasingularvectorcanonical}, and  
\textbf{(iii)} Mutual K-Nearest Neighbors (\textbf{Mutual-KNN})~\cite{huh2024platonicrepresentationhypothesis}.  
These metrics assess representational similarity across modalities, enabling us to investigate whether stronger explicit alignment leads to greater inter-modal consistency and improved downstream performance.

\subsubsection{Performance and Alignment in Redundancy-Dominant Scenarios} 
We first examine redundant information dominant scenarios, where redundancy is high and both unique and synergy are relatively low, as quantified by PID analysis (Table~\ref{tab:pid}) and synthetic setup.
\begin{enumerate}
    \item \noindent \textbf{Synthetic Dataset Results.}
    For high-redundancy synthetic datasets ($R = 6, 8$), we observe that unimodal encoder performance \textbf{monotonically improves and saturates} as alignment strength $\lambda$ increases (Figure~\ref{figure3}d, e). Alignment metrics also show a consistent upward trend (Figure~\ref{figure3}i, j), confirming the effectiveness of the explicit alignment objective in this setting.
    \item \noindent \textbf{Real-World Modalities with Minimal Unique Information.}
    Similar trends are observed in real-world scenarios where a modality contributes minimal unique information. In CMU-MOSEI (Vision in the Vision-Text pair), the Vision encoder exhibits negligible unique information ($U_1 = 0.001$) and moderate redundancy ($R = 0.123$). Its performance generally improves with increasing $\lambda$, peaking before a mild saturation (Figure~\ref{realworld}e, pink bars), while alignment metrics steadily rise (Figure~\ref{realworld}f). Likewise, in AV-MNIST (Audio in the Vision-Audio pair), the Audio encoder shows very low unique information ($U_2 = 0.03$). Its classification accuracy slightly increases as $\lambda$ grows, reaching a peak (Figure~\ref{realworld}a, blue bars), with alignment scores following a similar upward trend (Figure~\ref{realworld}b).
    \item \noindent \textbf{Analysis and Insights.} The consistent performance gains in both synthetic and real-world redundancy-dominant settings demonstrate that \textbf{explicit alignment is particularly effective when redundant information is existed and higher}. These findings empirically support the Platonic Convergence Hypothesis~\cite{huh2024platonic}, showing that maximizing inter-modal alignment is beneficial when an underlying shared structure exists.
 \noindent Notably, even within complex real-world datasets, individual modalities with minimal unique content can still benefit from alignment—provided redundancy exists. This directly answers our research question \textbf{\textit{RQ2}}, confirming that alignment benefits observed in controlled settings can transfer to practical, real-world scenarios. This highlights that alignment should not be applied solely based on coarse dataset-level labels, but rather guided by modality-level information decomposition. Strategic application of alignment, informed by fine-grained PID analysis, can thus maximize task performance by leveraging redundancy where present.
\end{enumerate}

\setlength{\tabcolsep}{1.5mm} 
\renewcommand{\arraystretch}{1.25} 
\begin{table}[t]
\centering
\small
\begin{tabular}{llcccc}
\hline
\textbf{Dataset} & \textbf{Modality} & \textbf{$R$} & \textbf{$U_1$} & \textbf{$U_2$} & \textbf{$S$} \\
\hline
\multirow{3}{*}{CMU-MOSEI} 
    & Vision-Text  & \textbf{0.123} & 0.001 & \textbf{0.163} & 0.005 \\
    & Vision-Audio & \textbf{0.116} & 0.010 & 0.001          & 0.012 \\
    & Audio-Text   & \textbf{0.127} & 0.001 & \textbf{0.248} & 0.002 \\
\hline
AV-MNIST & Vision-Audio & 0.170 & \textbf{0.970} & 0.030 & 0.040 \\
\hline
\multirow{3}{*}{MUSTARD} 
    & Vision-Text  & \textbf{0.150} & 0.020 & 0.010 & \textbf{0.340} \\
    & Vision-Audio & \textbf{0.140} & 0.020 & 0.010 & \textbf{0.200} \\
    & Audio-Text   & \textbf{0.160} & 0.010 & 0.010 & \textbf{0.370} \\
\hline
\end{tabular}
\caption{\small PID statistics for selected MultiBench datasets. Bold indicates the dominant information type within each modality pair.}
\label{tab:pid}
\end{table}

\subsubsection{Performance and Alignment in Uniqueness-Dominant Scenarios}We next examine scenarios where task-relevant information is primarily unique to individual modalities. This setting corresponds to low redundancy ($R$) and high uniqueness ($U$) as quantified by PID.

\begin{enumerate}
    \item \noindent \textbf{Synthetic Dataset Results.} For low-redundancy synthetic datasets ($R = 0, 2$), we observe a clear performance decline as $\lambda$ increases (Figure~\ref{realworld}c, d). Both unimodal encoders show a \textbf{monotonic decrease} in accuracy, indicating that explicit alignment hinders their ability to preserve distinct, modality-specific information. Notably, all alignment metrics still increase with $\lambda$ (Figure~\ref{figure3}f, g), confirming that the alignment objective is technically effective—even as it leads to degraded task performance.
    \item \noindent \textbf{Real-World Case.} A similar pattern emerges in the AV-MNIST dataset, where the vision encoder in the vision-audio pair is characterized by extremely high unique information ($U_1 = 0.97$). Starting from a strong baseline, its performance slightly declines as $\lambda$ increases, reaching a trough before a mild rebound at higher alignment strengths (Figure~\ref{realworld}a, pink bars). Meanwhile, alignment metrics continue to rise (Figure~\ref{realworld}b), further emphasizing the disconnect between alignment and performance in uniqueness-dominant conditions.
    \item \noindent \textbf{Analysis and Insights.}
    These results provide a clear answer to \textbf{\textit{RQ1}}, when task-relevant information is predominantly modality-specific, explicit alignment can be detrimental. By enforcing representational similarity, alignment suppresses the very distinctions that enable strong unimodal performance. In such cases, improvements in alignment scores do not translate into better task outcomes—and can even be inversely correlated.
    \noindent The behavior of the AV-MNIST Vision encoder offers real-world validation of this effect. Despite the dataset’s simplicity, the extreme dominance of unique visual information makes its performance particularly sensitive to forced alignment. These findings underscore the importance of tailoring alignment strategies to the underlying information structure of the data. In uniqueness-dominant settings, restraint in applying alignment is essential to preserve performance-critical modality-specific signals.
\end{enumerate}

\subsubsection{Performance and Alignment in Synergy-Dominant Scenarios}

We now investigate the impact of explicit alignment in scenarios dominated by synergistic information. Unlike redundancy or uniqueness, synergy reflects information that emerges only through the joint interpretation of modalities. This analysis focuses on the \textbf{MUSTARD dataset} (Vision-Audio pair), which our PID results (Table~\ref{tab:pid}) identify as \textbf{synergy-dominant} ($S = 0.20$), with relatively low redundancy ($R = 0.14$) and negligible unique information ($U_1 = 0.02$, $U_2 = 0.01$).

\begin{enumerate}
    \item \noindent \textbf{Observed Results.}
    As shown in Figure 5, increasing the alignment strength $\lambda$ leads to a general improvement in unimodal encoder performance. Specifically, the Vision encoder accuracy rises from 0.54 to 0.62, and the Audio encoder from 0.58 to 0.66 (Figure~\ref{realworld}c). In parallel, all alignment metrics increase monotonically with $\lambda$ (Figure~\ref{realworld}d), confirming that the explicit alignment objective is successfully promoting representational similarity.
    
    \item \noindent \textbf{Analysis and Insights.}
    These improvements offer a nuanced insight into alignment behavior in synergy-driven tasks. Although synergy, by definition, reflects information unavailable to any single modality, the observed performance gains suggest that explicit alignment can still be beneficial. We attribute this to the model's ability to leverage a \textbf{small but non-negligible amount of redundancy} ($R = 0.14$) present between the Vision and Audio modalities. By aligning representations around this shared signal, the model improves the robustness of its unimodal encoders—despite synergy being the dominant factor.
   \noindent However, it is important to note that the performance ceiling remains relatively low. This is expected, as sarcasm detection in MUSTARD depends heavily on \textbf{synergistic information}—content that emerges only through the interaction or fusion of both modalities and cannot be captured through alignment alone or only one encoder. While explicit alignment can enhance whatever redundant signals exist, it cannot uncover the deeper cross-modal dependencies that characterize synergy. Therefore, the utility of alignment in such settings is fundamentally constrained. These findings clarify that while modest gains are possible in synergy-dominant tasks, alignment remains insufficient for fully modeling tasks where the most critical signals reside beyond unimodal spaces.
\end{enumerate}

\begin{figure}[t]
    \centering
    \includegraphics[width=0.85\linewidth]{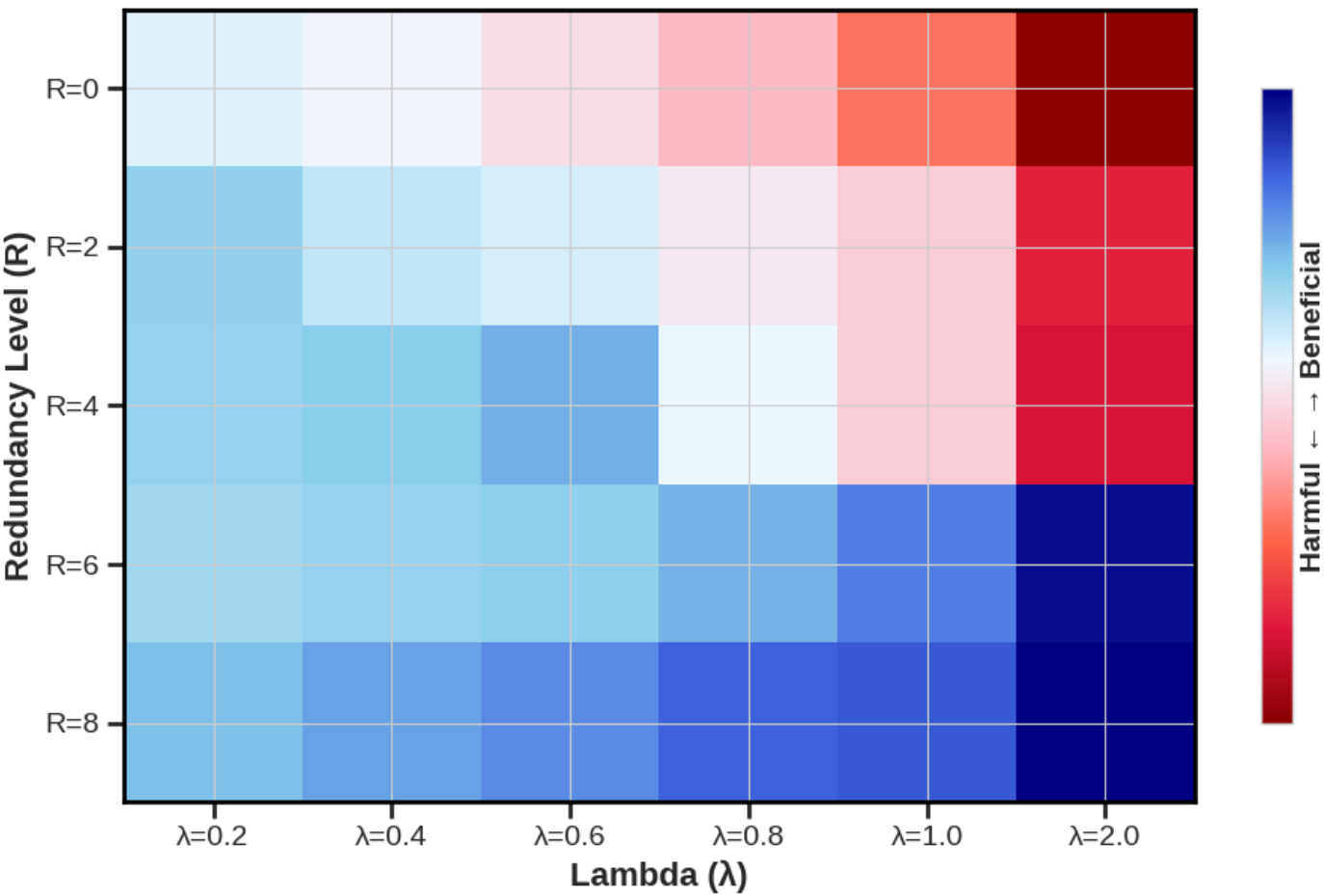}
   \caption{\small Effect of alignment strength $\lambda$ on synthetic-dataset encoder1 performance under varying redundancy.}
    \label{fig:enc1}
\end{figure}
\subsubsection{The Continuum of Alignment Utility: Towards Optimal Strategies}

We conclude by examining scenarios with mixed information distributions, where redundancy, uniqueness, and synergy coexist. These settings reveal that the utility of explicit alignment follows a \textbf{continuum}, often exhibiting an optimal alignment strength $\lambda^*$—a point beyond which further alignment becomes detrimental.

\begin{enumerate}
    \item \noindent \textbf{Synthetic Dataset Results.}
    In synthetic datasets with moderate redundancy ($R = 4$), unimodal encoder performance follows a characteristic \textbf{concave (inverted U-shaped)} trend (Figure~\ref{figure3}c). Performance initially improves with increasing $\lambda$, peaks around $\lambda = 0.4$–$0.6$, and then declines as over-alignment suppresses unique information. This trade-off is visually confirmed in Figure~\ref{fig:enc1}, where the transition from beneficial (blue) to harmful (red) impact is apparent at higher $\lambda$ values. Meanwhile, alignment metrics continue to increase monotonically (Figure~\ref{figure3}h), reinforcing the idea that stronger alignment does not necessarily imply better task performance.
 
    \item \noindent \textbf{Real-World Case: CMU-MOSEI Text Encoder.}
    The Text encoder from the Vision-Text pair in CMU-MOSEI offers a real-world illustration of this trend. With moderate redundancy ($R = 0.123$) and substantial unique information ($U_2 = 0.163$), it exhibits a similar inverted U-shape: performance increases with $\lambda$, peaks around $\lambda = 0.75$, and slightly drops thereafter (Figure~\ref{realworld}e, blue bars). Alignment metrics again increase steadily (Figure~\ref{realworld}f), highlighting the divergence between alignment strength and actual task utility.

    \item \noindent \textbf{Analysis and Insights.}
    These findings confirm that the effectiveness of explicit alignment is not binary, but depends on discovering an \textbf{optimal balance} between leveraging shared signals and preserving modality-specific information. While moderate alignment improves robustness by extracting redundancy, excessive alignment degrades performance by erasing valuable unique content. This validates our\textbf{\textit{ RQ1}} on the existence of an optimal $\lambda^*$, where alignment is most beneficial.
    \noindent The behavior of the CMU-MOSEI Text encoder exemplifies this balance: even as alignment improves, performance drops beyond $\lambda^*$, underscoring the importance of preserving distinct task-relevant signals. Overall, these results provide a principled guideline—\textbf{the optimal alignment strength should be tailored to the underlying information structure} of the multimodal data, offering practical insight for alignment-aware model design.
\end{enumerate}

\section{Conclusion}
\begin{figure}[t]
    \centering
    \includegraphics[width=0.95\linewidth]{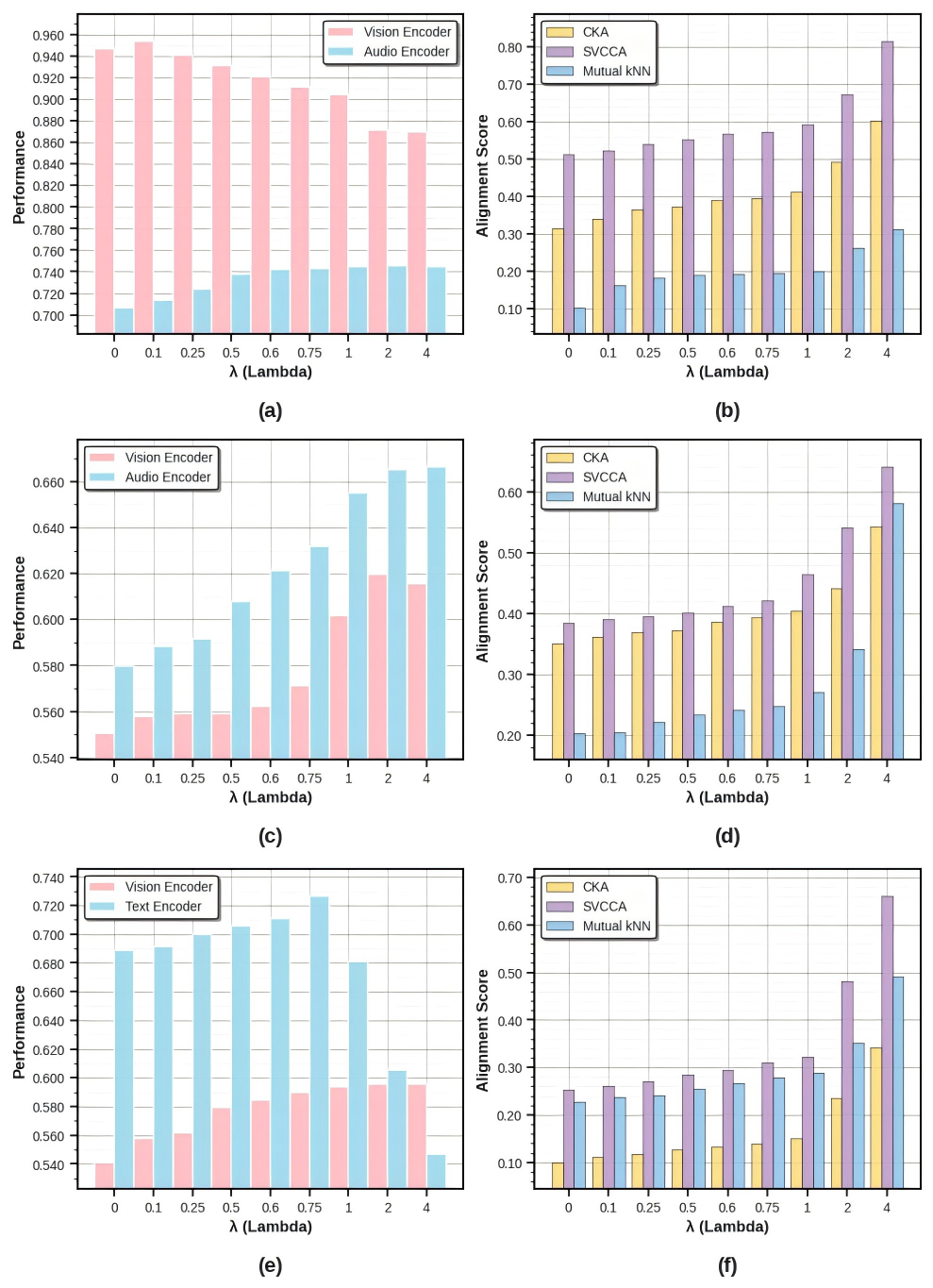}
\caption{\small  
Performance and representation alignment as a function of alignment strength $\lambda$ across different real-world datasets.  
(a)–(b): Results on AV-MNIST.  
(c)–(d): Results on MUStARD (V–A pair).  
(e)–(f): Results on CMU-MOSEI (V–T pair).
}
    \label{realworld}
\end{figure}

In this paper, we systematically examine when explicit cross-modal alignment helps or harms multimodal learning. 
We investigated the
link between explicitly enforced alignment and its impact on
both model performance and representation alignment under
the different underlying information structure of data. 
To achieve this, we propose a controlled experimental framework that integrates a tunable alignment loss with PID to analyze the relationship between alignment strength and information structure.
Our findings show that alignment is highly beneficial when modalities share redundant task-relevant information, but can be detrimental in uniqueness-dominant settings. We also find an optimal alignment strength in mixed-information regimes, suggesting that indiscriminate alignment may harm performance.

\section{Acknowledgments}
This research is supported by the National Research Foundation, Singapore under its National Large Language Models Funding Initiative (AISG Award No: AISG-NMLP-2024-001), NTU Start Up Grant and by the Ministry of Education, Singapore, under its Academic Research Fund Tier 1 (RG22/24). Any opinions, findings and conclusions or recommendations expressed in this material are those of the author(s) and do not reflect the views of National Research Foundation, Singapore. The computational work for this article was partially performed on resources of the National Supercomputing Centre (NSCC), Singapore (https://www.nscc.sg).

\bibliography{main.bib}

\clearpage
\section{1. Extended Related Work}
\label{app:related}
We provide a comprehensive review of prior work in multimodal alignment, covering both implicit emergence and contrastive learning approaches. This includes detailed discussion of the Platonic Representation Hypothesis, PID-based analysis, and recent advances in alignment-structure disentanglement.\\
\subsection{Multimodal Representation Alignment}
Recent research has shown that cross-modal alignment can emerge \textit{implicitly}~\cite{bonheme2022variationalautoencoderslearninsights, pmlr-v44-li15convergent, wang2018towards, lenc2015understanding}, even in the absence of any explicit optimization objective.  
This line of work~\cite{ziyin2025proofperfectplatonicrepresentation}, exemplified by the \textit{Platonic Representation Hypothesis (PRH)}~\cite{huh2024platonicrepresentationhypothesis}, suggests that as unimodal models scale in data, architecture, and capacity, their internal representations exhibit natural convergence—capturing a shared statistical structure of the underlying reality.  
Empirical studies using representational similarity metrics such as CKA~\cite{kornblith2019similarityneuralnetworkrepresentations} and SVCCA~\cite{raghu2017svccasingularvectorcanonical} further demonstrate that such emergent alignment is positively correlated with downstream task performance.
However, these studies remain largely \textit{observational}: they do not manipulate alignment strength or assess its causal effect on model behavior.  
Notably, Tjandrasuwita et al.~\cite{tjandrasuwita2025understandingemergencemultimodalrepresentation} take a step in this direction by analyzing alignment–performance relationships conditioned on the information structure of the data.  
Yet, their approach still focuses on naturally occurring alignment, without systematically intervening in or controlling alignment during training.
In contrast, our work explicitly enforces and varies the degree of alignment between modality-specific encoders.  
We analyze how this controlled alignment affects both \textit{unimodal task performance} and \textit{cross-modal representation similarity}, under varying information structures.  
This allows us to move beyond observation and toward a understanding of when alignment helps—or hurts—unimodal learning.   \subsection{Cross-modal Alignment via Contrastive Learning}
Contrastive learning has emerged as a dominant paradigm for \textit{explicitly enforcing} cross-modal alignment.  
Methods such as CLIP~\cite{radford2021learning}, VideoCLIP~\cite{xu2021videoclip}, and CrossCLR~\cite{zolfaghari2021crossclr} employ contrastive objectives to map representations from different modalities into a shared embedding space.  
More recent work, such as Dufumier et al.~\cite{dufumier2025alignmultimodalcontrastivelearning}, questions \textit{what} aspects should be aligned and proposes refined objectives to better preserve task-relevant, modality-specific information.
However, most existing contrastive methods operate under a \textit{universal alignment assumption}—that maximizing cross-modal similarity is universally beneficial~\cite{cai2025valuecrossmodalmisalignmentmultimodal, zong2024self, jiang2023visionlanguagepretrainingcontrastive, jiang2023understanding}.  
In doing so, they often overlook the underlying structure of multimodal signals, including the distribution of \textit{redundant, unique, and synergistic} information.  
As a result, strong alignment constraints may over-regularize the representations, suppressing important modality-specific features and degrading unimodal performance.
To address this gap, we introduce a \textit{controllable contrastive learning module} that allows explicit modulation of alignment strength. 
By systematically varying this alignment factor, we are able to analyze when and how cross-modal alignment benefits or harms unimodal encoder performance.  
Leveraging both synthetic datasets with controlled information structures and real-world datasets analyzed through \textit{Partial Information Decomposition (PID)}, our framework offers actionable insights for designing alignment-aware multimodal learning systems.

\section{2: Alignment Computation}
To evaluate how well the representations from different modalities align in a shared latent space, we employ three complementary alignment metrics: \textit{Centered Kernel Alignment (CKA)}, \textit{Singular Vector Canonical Correlation Analysis (SVCCA)}, and \textit{Mutual-KNN Alignment}. Each of these metrics captures distinct perspectives of alignment: global similarity, canonical subspace agreement, and local structural consistency, respectively.

Let $F_A, F_B \in \mathbb{R}^{n \times d}$ denote the hidden representations extracted from $n$ samples by two unimodal encoders (e.g., vision and text). All representations are mean-centered before alignment computation unless otherwise specified.

\subsection*{Section 2.1: Centered Kernel Alignment}

\textbf{Motivation.}  
CKA~\cite{kornblith2019similarity} is a widely used metric for measuring the similarity between two sets of representations. It is invariant to isotropic scaling and orthogonal transformation, making it suitable for comparing features across heterogeneous modalities. In this work, CKA serves as a global alignment metric to assess the extent to which entire representation spaces of two modalities are linearly aligned.

\textbf{Method.}  
We first compute the Hilbert-Schmidt Independence Criterion~\cite{gretton2005measuring} (HSIC) using a linear kernel:

\begin{equation}
\text{HSIC}(F_A, F_B) = \frac{1}{(n - 1)^2} \left\| F_A^\top F_B \right\|_F^2
\end{equation}

CKA is then defined by normalizing the HSIC:

\begin{equation}
\text{CKA}(F_A, F_B) = \frac{\text{HSIC}(F_A, F_B)}{\sqrt{\text{HSIC}(F_A, F_A) \cdot \text{HSIC}(F_B, F_B)}}
\end{equation}

\textbf{Usage.}  
In our experiments, we use CKA to compare representation similarity across alignment strength $\lambda$ in both synthetic and real-world datasets. A rising CKA value with increasing $\lambda$ indicates stronger global alignment between modalities.

\subsection*{Section 2.2: Singular Vector Canonical Correlation Analysis (SVCCA)}

\textbf{Motivation.}  
SVCCA is a two-step method that first reduces each representation to its most informative subspace via SVD, then evaluates correlation between those subspaces using canonical correlation analysis~\cite{hardoon2004canonical} (CCA). This method focuses on subspace-level alignment and is sensitive to dominant directions of variation.

\textbf{Method.}  
Given $F_A, F_B \in \mathbb{R}^{n \times d}$:

\begin{enumerate}
    \item \textbf{Preprocessing:}  
    Normalize each feature dimension:
    \[
        \hat{F}_i = \frac{F_i - \mu(F_i)}{\sigma(F_i) + \epsilon}, \quad i \in \{A, B\}
    \]
    
    \item \textbf{Dimensionality Reduction:}  
    Perform truncated SVD on each normalized matrix to extract the top-$k$ left singular vectors:
    \[
        U_A, \_, \_ = \text{SVD}_k(\hat{F}_A), \quad U_B, \_, \_ = \text{SVD}_k(\hat{F}_B)
    \]

    \item \textbf{Canonical Correlation:}  
    Apply CCA to $U_A$ and $U_B$:
    \[
        (U'_A, U'_B) = \text{CCA}(U_A, U_B)
    \]

    \item \textbf{Final Similarity:}  
    Compute the average Pearson correlation of corresponding dimensions:
    \[
        \text{SVCCA}(F_A, F_B) = \frac{1}{k} \sum_{j=1}^{k} \rho(U'_A[:, j], U'_B[:, j])
    \]
\end{enumerate}

\textbf{Usage.}  
We use SVCCA to quantify alignment between high-variance directions across modalities. It complements CKA by focusing on aligned canonical axes rather than full-space alignment.

\subsection*{Section A3: Mutual-KNN Alignment}

\textbf{Motivation.}  
Global metrics like CKA and SVCCA may overlook local structural consistency. To address this, we compute alignment based on shared local neighborhoods using Mutual-KNN~\cite{huh2024platonicrepresentationhypothesis}. This metric captures how similarly two modalities structure their representations around each data point.

\textbf{Method.}  
Let $f_{A}^{(i)}$ and $f_{B}^{(i)}$ denote the $i$-th row vectors in $F_A$ and $F_B$. Let $\text{knn}(f_{A}^{(i)})$ denote the $k$ nearest neighbors of $f_{A}^{(i)}$ in modality $A$.

We define pairwise neighborhood agreement indicator:

\begin{equation}
\alpha(i, j) = \mathbb{1}\left\{ f_{A}^{(j)} \in \text{knn}(f_{A}^{(i)}) \land f_{B}^{(j)} \in \text{knn}(f_{B}^{(i)}) \land i \neq j \right\}
\end{equation}

The total mutual alignment score is:

\begin{equation}
\text{Align}_{\text{MKNN}}(F_A, F_B) = \sum_{i=1}^{n} \sum_{j=1}^{n} \alpha(i, j)
\end{equation}

To normalize and obtain a scale-invariant score, we compute:

\begin{equation}
\text{MKNN}(F_A, F_B) = \frac{\text{Align}_{\text{MKNN}}(F_A, F_B)}{\sqrt{\text{Align}_{\text{MKNN}}(F_A, F_A) \cdot \text{Align}_{\text{MKNN}}(F_B, F_B)}}
\end{equation}

\textbf{Usage.}  
Mutual-KNN is particularly informative when the representations are well-structured locally (e.g., when clusters exist). It provides a finer-grained view of alignment effectiveness, especially in low-redundancy or synergy-rich conditions.

\section{3: Dataset Details}
\subsection{Section 3.1: AV-MNIST Dataset}
The AV-MNIST~\cite{perez2019mfas} dataset is a synthetic audio-visual benchmark created by pairing spoken digits from the Free Spoken Digit Dataset (FSDD) with handwritten digits from the MNIST image dataset~\cite{lecun2002gradient}. The task is a 10-class classification problem, where each input pair corresponds to a digit from 0 to 9. For each digit identity, the corresponding image and audio are matched to ensure label consistency across modalities.
While previous works~\cite{perez2019mfas} often introduce artificial difficulty by discarding visual information via PCA (e.g., preserving only 25\% of variance) and injecting environmental noise (e.g., ESC-50) into the audio, we retain both modalities in their clean form. This design choice is intentional: it ensures that the audio and visual signals contain overlapping task-relevant information, allowing us to explore redundancy-dominant alignment scenarios in a controlled fashion.

The dataset is split into 55,000 training samples, 5,000 validation samples, and 10,000 test samples, maintaining a uniform label distribution across all splits. For each sample, the MNIST image and the corresponding audio waveform represent the same digit class.

In our experiments, we train a Vision Transformer (ViT) on the MNIST images. Each $28 \times 28$ grayscale image is divided into $4 \times 4$ patches, resulting in a sequence length of 49 (including a learnable \texttt{[CLS]} token). For the audio modality, we preprocess raw FSDD waveforms using the \textit{librosa} library to extract 36-dimensional Mel-Frequency Cepstral Coefficients (MFCCs)~\cite{mcfee2015librosa}. The MFCC sequences are zero-padded or truncated to a fixed maximum length of 20, resulting in a consistent input shape for the audio Transformer encoder.

This setup allows us to analyze how explicit alignment affects unimodal performance when modalities exhibit high redundancy and minimal uniqueness—making AV-MNIST a suitable testbed for studying alignment in redundancy and uniqueness scenario.

\subsection{Section 3.2: MUStARD Dataset}
The MUStARD~\cite{castro2019towards} (Multimodal Sarcasm Detection) dataset is a benchmark multimodal resource designed for the task of automatic sarcasm detection in conversational video settings. It consists of short video clips sampled from popular American TV shows such as \textit{Friends}, \textit{The Golden Girls}, \textit{The Big Bang Theory}, and \textit{Sarcastic Anonymous}. Each data sample corresponds to an utterance that has been manually annotated as \textit{sarcastic} or \textit{non-sarcastic}.
Each utterance is aligned with three distinct modalities:
\begin{itemize}
    \item \textbf{Visual:} A sequence of video frames capturing facial expressions, gestures, and contextual cues from the speaker.
    \item \textbf{Audio:} The corresponding speech waveform, encoding vocal tone, prosody, and other paralinguistic features.
    \item \textbf{Text:} The transcribed utterance content.
\end{itemize}

The multimodal nature of sarcasm presents a unique challenge: neither unimodal textual nor visual/auditory cues alone are often sufficient for reliable prediction. Instead, effective sarcasm detection in MUStARD typically relies on synergistic integration of contextual signals across modalities—making it an ideal testbed for evaluating alignment strategies in synergy-dominant conditions.

The dataset contains a total of 690 labeled instances, with a predefined split of 414 samples for training, 138 for validation, and 138 for testing. The dataset is moderately imbalanced with respect to class distribution and is known for its relatively small size and rich semantic complexity. It has been widely adopted in multimodal sentiment and emotion analysis literature, particularly for evaluating fusion techniques and modeling subtle interpersonal communication signals.
In our experiments, we use the pre-extracted features for vision, audio, and text modalities as provided by the MultiBench suite \cite{liang2021multibenchmultiscalebenchmarksmultimodal}. All modalities are temporally aligned at the utterance level and undergo standardized preprocessing before input to the encoder networks.

\subsection{Section 3.3: CMU-MOSEI Dataset}
The CMU-MOSEI~\cite{zadeh2018multimodal} (Multimodal Opinion Sentiment and Emotion Intensity) dataset is one of the largest and most widely used benchmarks for multimodal sentiment and emotion analysis. It contains over 23,500 sentence-level utterances, manually segmented from YouTube monologue videos featuring more than 1,000 distinct speakers. The dataset is gender-balanced and covers a diverse range of topics, ensuring broad generalization across speaker styles and content domains.
Each data sample is composed of:
\begin{itemize}
    \item \textbf{Visual modality:} A sequence of facial visual frames capturing fine-grained gestures and expressions.
    \item \textbf{Audio modality:} Raw speech waveforms containing prosodic cues and vocal dynamics.
    \item \textbf{Text modality:} Accurately transcribed and punctuated sentences aligned with the video and audio segments.
\end{itemize}
The dataset is divided into 16,265 training samples, 1,869 validation samples, and 4,643 test samples, for a total of 22,777 instances. All utterances are temporally aligned across modalities and are accompanied by continuous-valued sentiment intensity labels ranging from $-3$ (strongly negative) to $+3$ (strongly positive).
In our experiments, we formulate sentiment prediction as a binary classification task by thresholding the original regression labels into two classes (positive vs. negative). We train our models using the L1 loss on the original continuous labels and report binary classification performance. This hybrid objective allows us to retain fine-grained learning signals while aligning with standard evaluation metrics in the literature.
The CMU-MOSEI dataset exhibits a rich mix of redundant and unique information across modalities. As shown in our PID analysis, the text modality tends to provide substantial unique task-relevant information, while the visual and audio modalities often exhibit moderate redundancy. This diverse information structure makes CMU-MOSEI a valuable benchmark for evaluating the conditional effects of explicit alignment across modality pairs.

\subsection{Section 3.4: Synthetic Dataset}
The synthetic dataset~\cite{tjandrasuwita2025understandingemergencemultimodalrepresentation} used in this work is designed to explore the impact of different information relationships—such as redundancy and uniqueness—between modalities on the performance of alignment methods. The dataset is composed of two modalities, each represented by a 12-dimensional feature space, with the goal of simulating real-world multimodal tasks where both redundant and unique information is present across modalities.

Each sample in the dataset consists of:
\begin{itemize}
    \item \textbf{Modality 1 ($X_1$):} A set of binary features representing modality-specific information. This set contains both shared features (redundant with the other modality) and unique features (exclusive to modality 1).
    \item \textbf{Modality ($X_2$):} A corresponding set of binary features, similarly containing shared features (redundant with modality 1) and unique features (exclusive to modality 2).
\end{itemize}

\textbf{Data Generation Process:}  
The data is generated using a structured framework that controls the amount of overlap (redundancy) between the two modalities. The key parameters involved in the generation process include:
\begin{itemize}
    \item \textbf{Redundancy (R):} The number of shared features between the two modalities. \( R \in \{0, 1, 2, \dots, 8 \} \).
    \item \textbf{Uniqueness (U):} The amount of modality-specific information. \( U = F - R \), where \( F = 8 \) is the total number of features per modality.
    \item \textbf{Noise Features:} Irrelevant features are added to the dataset to simulate real-world conditions, ensuring that some features do not contribute to the label generation.
\end{itemize}

The dataset is designed to generate tasks with varying levels of mutual information and redundancy, creating a range of complexity for alignment analysis. The feature allocation for each redundancy level \( R \) follows these rules:
\begin{itemize}
    \item Shared Features: \( R \) features are randomly selected from an 8-dimensional shared pool.
    \item Unique Features for Modality ($X_1$): \( \lfloor U/2 \rfloor \) features are chosen from a pool of 4 unique features.
    \item Unique Features for Modality ($X_2$): \( \lceil U/2 \rceil \) features are chosen from another pool of 4 unique features.
\end{itemize}

This structured feature space allows us to control and vary the amount of shared and unique information, offering a controlled environment to study how alignment methods behave under different data conditions.

\textbf{Label Generation:}  
Labels for the dataset are generated using a temperature-controlled softmax function that reflects task difficulty:
\[
    p(y | x_1, x_2) = \text{softmax}\left(W \cdot [x_c^{(R)}, x_{u1}^{(U_1)}, x_{u2}^{(U_2)}] / \tau \right)
\]
where:
\begin{itemize}
    \item \( W \in \mathbf{R}^{8 \times 4} \) is a randomly initialized weight matrix.
    \item \( \tau \) is the temperature parameter that controls task complexity by modulating mutual information between modalities.
    \item \( x_c^{(R)}, x_{u1}^{(U_1)}, x_{u2}^{(U_2)} \) are the selected shared and unique features based on the redundancy configuration.
\end{itemize}

\textbf{Dataset Scale and Splits:}  
The dataset is split into training, validation, and test sets, ensuring a balanced class distribution across each split:
\begin{itemize}
    \item \textbf{Training samples:} 45,920 (70\%)
    \item \textbf{Validation samples:} 9,828 (15\%)
    \item \textbf{Test samples:} 9,828 (15\%)
    \item \textbf{Total samples:} 65,536 per configuration
\end{itemize}

\textbf{Configuration Matrix:}  
The experimental design spans:
\begin{itemize}
    \item \textbf{Redundancy levels:} 9 values \( R \in \{0, 1, \dots, 8 \} \)
    \item \textbf{Mutual information levels:} Multiple values of \( \tau \), enabling the creation of datasets with different levels of mutual information.
    \item \textbf{Random seeds:} 3 random seeds for feature generation and weight initialization.
\end{itemize}

This results in multiple synthetic datasets that enable comprehensive analysis of alignment emergence and unimodal performance across various redundancy and uniqueness configurations.

\textbf{Input Properties:}
\begin{itemize}
    \item \textbf{Dimensionality:} 12D per modality, consistent across all configurations.
    \item \textbf{Feature type:} Binary {0,1} features with deterministic generation.
    \item \textbf{Sequence length:} Single time-step (static features).
    \item \textbf{Missing data:} No missing data; all samples are complete and paired.
\end{itemize}

\textbf{Task Properties:}
\begin{itemize}
    \item \textbf{Task type:} 4-class classification task.
    \item \textbf{Class balance:} Uniform distribution across all configurations.
    \item \textbf{Decision boundaries:} Smooth, learnable relationships via softmax generation.
    \item \textbf{Noise level:} Controlled via the temperature parameter \( \tau \) and inclusion of irrelevant features.
\end{itemize}

\section{4: Experimental Details}
\subsection{Section 4.1: Synthetic Dataset Experiments.} 
To systematically investigate how explicit cross-modal alignment affects unimodal encoder performance under varying information structures, we adopt a controlled experimental setup using synthetic datasets. We train two Multi-Layer Perceptrons (MLPs)—each corresponding to one modality—with identical architectures to ensure a fair comparison. Both MLPs are implemented with a hidden size of 12, which matches the input feature dimension (comprising 8 shared and 4 unique dimensions). The network depth is fixed at 3 layers to avoid unnecessary complexity while preserving expressive capacity.

Training is conducted using the AdamW optimizer, chosen for its robustness in handling sparse gradients and improved generalization. We perform extensive grid search over learning rates $\{0, 10^{-1}, 10^{-2}, 10^{-3}, 10^{-4}, 10^{-5}\}$ and weight decays from the same set, aiming to ensure optimal convergence across varying degrees of alignment strength. The alignment loss is computed over mini-batches of size 512, which we empirically found to provide a stable estimate of similarity metrics without inducing overfitting. To ensure statistical robustness, all experiments are repeated with 4 different random seeds, and we report the mean results across these runs.

\subsection{Section 4.2: Real-World Dataset Experiments.} 
To validate and generalize findings from the synthetic domain, we extend our experiments to three real-world multimodal datasets. For the CMU-MOSEI and MUStARD affective computing datasets, we utilize pre-extracted features from video, audio, and text modalities. For AV-MNIST, we separately process image and audio inputs, training a dedicated Vision Transformer (ViT) for image sequences and a Transformer encoder for the audio modality.

Each Transformer encoder utilizes a single-head self-attention mechanism to maintain architectural simplicity. The embedding dimension of each encoder is set equal to the corresponding input feature dimension, avoiding unnecessary projection prior to alignment. For classification tasks, we append a learnable \textit{[CLS]} token to the input sequence; its hidden state is used as the global sequence representation. This representation is then used to compute inter-layer alignment when applicable. In scenarios where \textit{[CLS]} is not used, we instead apply average pooling over the sequence.

All real-world models are optimized using AdamW. For each dataset, we conduct a hyperparameter search over learning rates $\{10^{-3}, 5 \times 10^{-4}, 10^{-4}, 5 \times 10^{-5}, 10^{-5}\}$ and weight decays $\{0, 10^{-1}, 10^{-2}, 10^{-3}, 10^{-4}, 10^{-5}\}$ across different model depths, ensuring that model capacity is adapted to the complexity of the data.

\subsection{Section 4.3: Projection Head Design.} 
In the synthetic dataset setting, both modality encoders share identical output dimensionality, eliminating the need for additional projection. In contrast, real-world datasets often involve heterogeneous modalities (e.g., image, audio, and text) with incompatible embedding sizes. To facilitate alignment under such settings, we introduce dedicated MLP-based projection heads to map the encoder outputs into a shared latent space.

These projection heads are trained jointly with the encoders using a contrastive loss. We tune the projection learning rate over $\{10^{-4}, 5 \times 10^{-4}, 10^{-3}, 5 \times 10^{-3}, 10^{-2}\}$ and apply weight decay from the set $\{0, 10^{-5}, 10^{-4}, 10^{-3}, 10^{-2}\}$ to ensure proper regularization. To explore the dimensionality of the alignment space, we experiment with projection sizes of $\{64, 128, 256\}$.

Recognizing the differing stability and noise levels across modalities, we adopt modality-specific dropout strategies. For relatively clean and semantically stable modalities such as vision and text, we apply low dropout rates. For more variable or noisy modalities—particularly audio—we apply higher dropout to prevent overfitting and improve generalization. The dropout rate is selected from $\{0.0, 0.01, 0.02, 0.05, 0.1, 0.15, 0.2\}$ based on validation performance.

\subsection{Section 4.4: Implementation Notes}

All model implementations, including the training of encoders, projection heads, and contrastive alignment loss, are carried out on a 2* RTX 6000 Ada GPU workstation. We provide modular configurations, ensuring that the training protocols can be consistently reused across both synthetic and real-world datasets. 

To facilitate reproducibility, all hyperparameter choices and configurations are documented in detail. Importantly, all experimental results reported in this work are based on the average performance across 5 different random seeds. This approach helps to mitigate any potential biases arising from specific initializations, ensuring that the reported findings are robust and representative of general model behavior.

\section{5: Additional Experiment Result} 
In the main text, we presented experimental results for one modality pair from both the MOSEI and MUSTARD datasets. To provide a comprehensive view of the performance and alignment across different modality pairs, we now include the remaining two modality pair results (Figure 6 and Figure 7) in this appendix. Specifically, we present additional experiments on the MOSEI dataset (Vision-Audio pair) and the MUSTARD dataset (Vision-Text and Audio-Text pairs). These results further demonstrate the impact of modality-specific alignments and redundancy-uniqueness trade-offs on model performance across various configurations of \( \lambda \).

\begin{figure}[h]
    \centering
    \includegraphics[width=\linewidth]{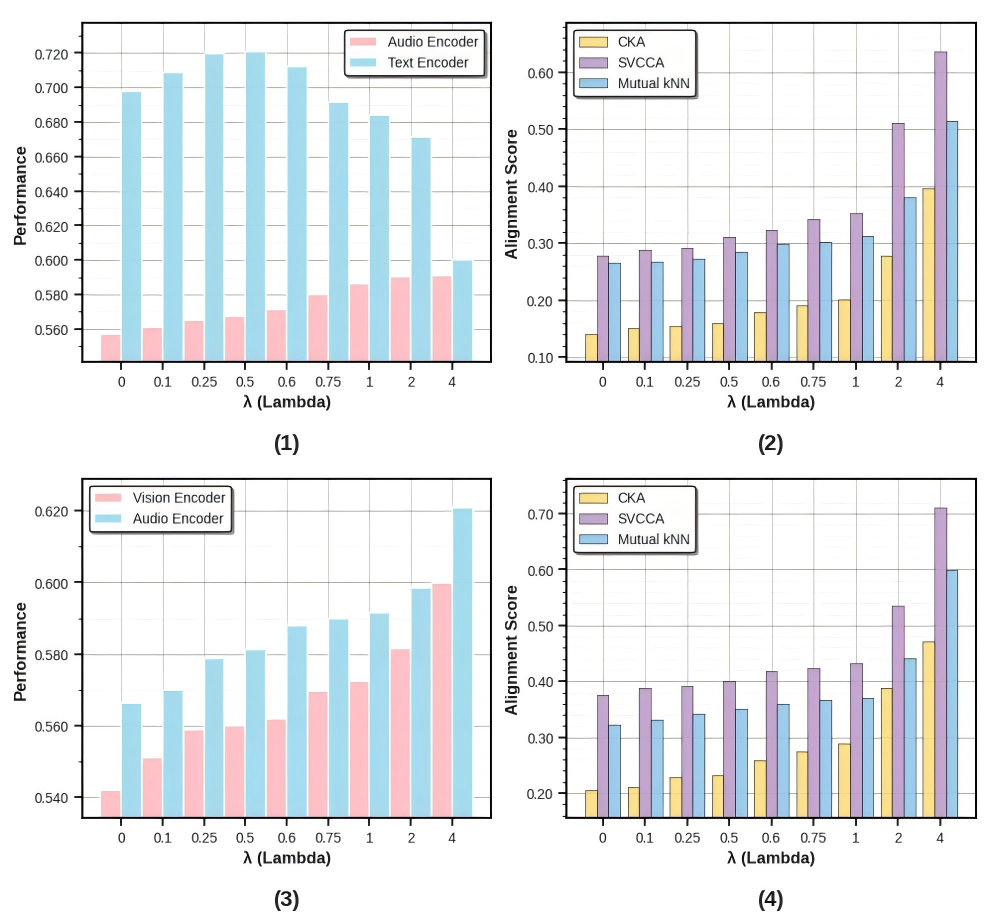}
   \caption{Performance and Alignment Metrics vs $\lambda$ for Different real-world Datasets. (1) \&  (2): experiment results on the MOSEI dataset (Audio-Text pair). (3) \& (4): experiment results on the MOSEI dataset (Vision-Audio pair).
}
    \label{app}
\end{figure}

\begin{figure}[h]
    \centering
    \includegraphics[width=\linewidth]{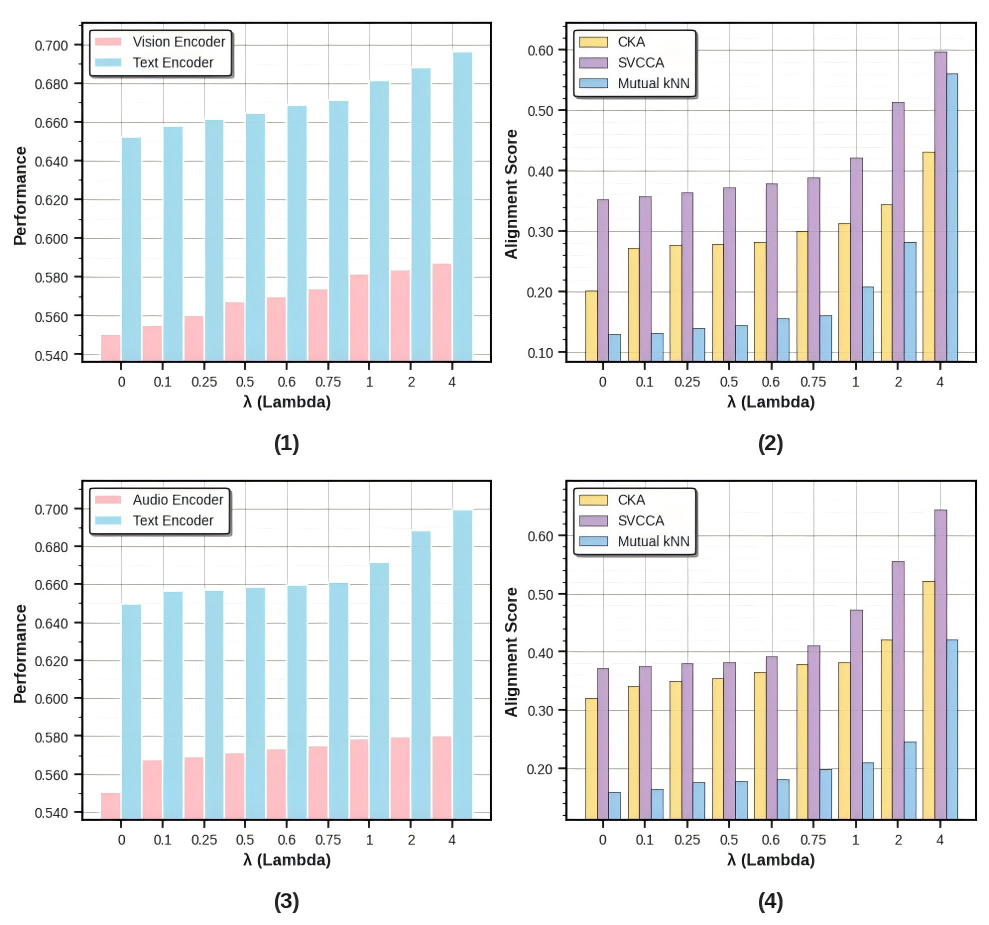}
   \caption{Performance and Alignment Metrics vs $\lambda$ for Different real-world Datasets. (1) \&  (2): experiment results on the MUSTARD dataset (Vision-Text pair). (3) \& (4): experiment results on the MUSTARD dataset (Audio-Text pair).
}

    \label{app}
\end{figure}

\end{document}